\documentclass[letterpaper, 10 pt, conference]{ieeeconf}

\IEEEoverridecommandlockouts
\usepackage{amsmath}
\usepackage[dvipsnames]{xcolor}
\usepackage{amsfonts}
\usepackage{algorithm}
\usepackage{graphicx}
\usepackage{caption}
\usepackage{subcaption}
\usepackage{amssymb}
\usepackage{hyperref}
\usepackage{cite}
\usepackage{pifont}
\usepackage{algpseudocode}

\renewcommand{\baselinestretch}{0.94}

\title{\LARGE \bf
Stage-Wise Reward Shaping for Acrobatic Robots: \\
A Constrained Multi-Objective Reinforcement Learning Approach 
}

\author{Dohyeong Kim$^{1*}$, Hyeokjin Kwon$^{2*}$, Junseok Kim$^{1}$, Gunmin Lee$^{1}$, and Songhwai Oh$^{1,2}$ 
\thanks{$^{1}$D. Kim, J. Kim, G. Lee, and S. Oh are with the Department of Electrical and Computer Engineering and ASRI, Seoul National University, Seoul 08826, Korea (e-mail: \{dohyeong.kim, junseok.kim, gunmin.lee\}@rllab.snu.ac.kr, songhwai@snu.ac.kr).
$^{2}$H. Kwon ans S. Oh are with the Interdisciplinary Program in Artificial Intelligence and ASRI, Seoul National University, Seoul 08826, Korea (e-mail: hyeokjin.kwon@rllab.snu.ac.kr)}%
}

\begin{document}

\makeatletter
\g@addto@macro\@maketitle{
    \begin{figure}[H]
    \setlength{\linewidth}{\textwidth}
    \setlength{\hsize}{\textwidth}
    \centering
    \begin{subfigure}[b]{1.0\linewidth}
        \centering    
        \includegraphics[width=1\linewidth]{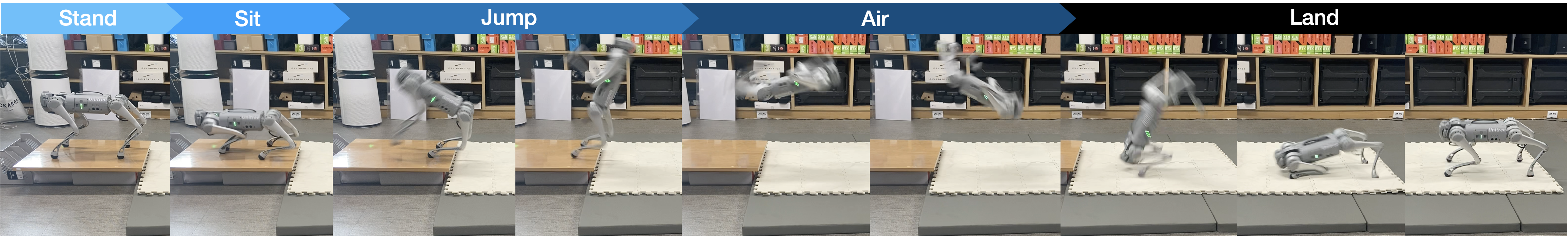}
    \end{subfigure}
    \hfill
    \vspace{0.1pt}
    \begin{subfigure}[b]{1.0\linewidth}
        \centering    
        \includegraphics[width=1\linewidth]{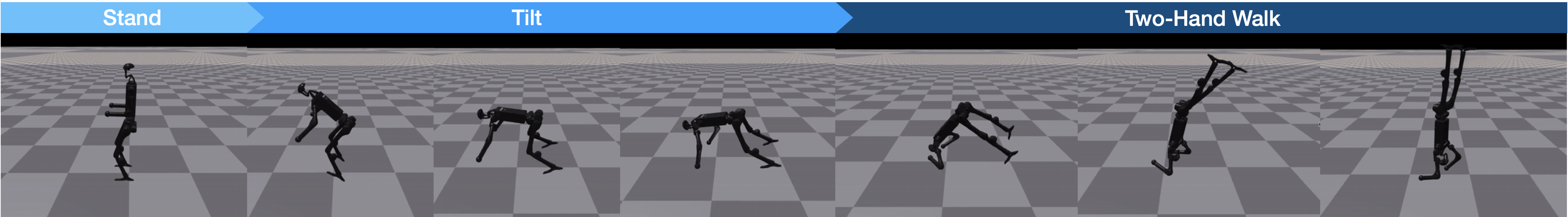}
    \end{subfigure}
    \caption{\textbf{Snapshots of robot movements with stage transitions.}
The top row shows a quadrupedal robot performing a back-flip in a real-world environment, and the bottom row shows a humanoid performing a two-hand walk in simulation. Each sequence of snapshots is annotated at the top with the current stage of the episode.}
    \label{fig: snapshots}
    \vspace{-10pt}
    \end{figure}
}
\makeatother

\maketitle
\thispagestyle{empty}
\pagestyle{empty}

\setcounter{figure}{1}


\begin{abstract}
As the complexity of tasks addressed through reinforcement learning (RL) increases, the definition of reward functions also has become highly complicated. 
We introduce an RL method aimed at simplifying the reward-shaping process through intuitive strategies. 
Initially, instead of a single reward function composed of various terms, we define multiple reward and cost functions within a constrained multi-objective RL (CMORL) framework. 
For tasks involving sequential complex movements, we segment the task into distinct stages and define multiple rewards and costs for each stage.
Finally, we introduce a practical CMORL algorithm that maximizes objectives based on these rewards while satisfying constraints defined by the costs. 
The proposed method has been successfully demonstrated across a variety of acrobatic tasks in both simulation and real-world environments. Additionally, it has been shown to successfully perform tasks compared to existing RL and constrained RL algorithms.
Our code is available at \url{https://github.com/rllab-snu/Stage-Wise-CMORL}.
\end{abstract}


\section{INTRODUCTION}

Recently, reinforcement learning (RL) has driven significant progress in real-world robotic applications \cite{lee2020learning, margolis2023walk, zhang2023learning, kumar2021rma, smith2022legged}. 
Quadrupedal robots have demonstrated stable locomotion on rough terrain \cite{lee2020learning, ji2022concurrent} and performed parkour-like stunts \cite{caluwaerts2023barkour, cheng2024extreme}, while bipedal robots have successfully tackled challenging tasks such as jumping and running \cite{seikmann2021stairs, li2023cassie}.
In order to accomplish such legged robot tasks, a reward function should be defined considering multiple factors, such as task performance, safety, and energy efficiency.
Consequently, reward functions are generally formalized as a sum of various terms related to performance, safety, and regularization \cite{lee2020learning, ji2022concurrent, margolis2023walk}.
However, due to the numerous terms, the reward-shaping process, defining each term and its respective weight, can be laborious and challenging. 
Simplifying this process is essential to apply RL to a broader range of tasks.

In the case of acrobatic tasks involving complex movements, such as rolls and back-flips, the difficulty of designing rewards increases significantly. 
By taking the back-flip as an example, this task requires a focus on jumping at the beginning of the episode and on landing after the jump.
As a result, the proportions of reward terms related to jumping and landing should be adjusted dynamically, complicating the reward-shaping process.
Alternatively, imitation RL methods using motion capture data or animatronic data have been developed \cite{peng2018deepmimic, peng2022ase, grandia2024bipdal}, wherein the reward is defined to minimize the pose difference between the robot and the collected data.
However, these methods are expensive as they require collecting extensive data for each task. 
Therefore, a method is required that can intuitively design reward functions without relying on additional imitation data.

\begin{figure}[!tb]
    \centering
    \includegraphics[width=1\linewidth]{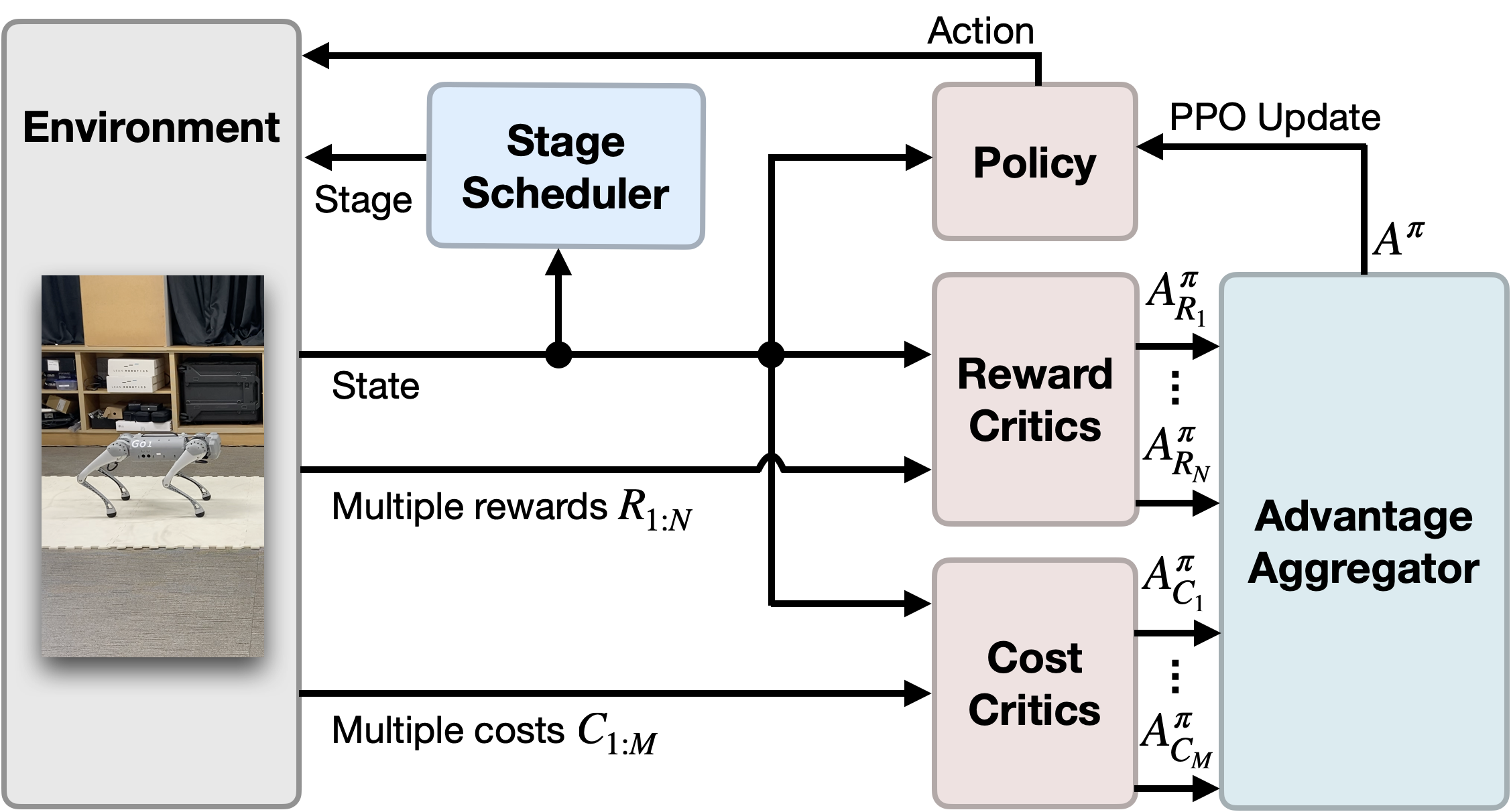}
    \caption{\textbf{Overview of the proposed framework.} 
    An environment provides multiple rewards and costs, and critics compute value estimates for each reward and cost. 
    These estimates are aggregated to calculate the overall advantage, as detailed in Sec. \ref{sec: policy update}, which is subsequently used for policy updates. Additionally, a stage scheduler updates the current stage based on a user-defined rule for stage transitions.}
    \label{fig: overview}
    \vspace{-10pt}
\end{figure}

In this paper, we propose an RL method that defines multiple reward functions in a stage-wise manner by utilizing a constrained multi-objective RL (CMORL) framework \cite{kim2024scale}.
Examples of the stages and representative results are presented in Fig. \ref{fig: snapshots}. 
To simplify the reward-shaping process, our approach does not integrate multiple terms into a single reward. 
Instead, each term is treated as an independent reward or cost function within the CMORL framework, where the cost functions correspond to safety-related terms, such as body collision and joint limit.
In addition, to facilitate reward shaping for acrobatic tasks requiring a series of complex movements, we propose segmenting tasks into several stages and defining reward and cost functions for each stage. 
The overall framework is presented in Fig. \ref{fig: overview}, and examples of the reward and cost definitions are provided in Table \ref{tab: example of reward}.
In this framework, a new variant of proximal policy optimization (PPO) \cite{schulman2017proximal} adapted for CMORL, named \emph{constrained multi-objective PPO (CoMOPPO)}, is introduced for policy updates. 
Also, to ensure successful real-world deployment, sim-to-real techniques, such as domain randomization \cite{cheng2024extreme, zhang2023learning} and teacher-student distillation \cite{lee2020learning}, are employed.

The proposed method has been applied to a range of acrobatic tasks on quadrupedal and humanoid robots: \emph{two-hand walks, back-flips, side-flips}, and \emph{side-rolls}.
For real-world evaluation, the side-roll, back-flip, and two-hand walk tasks were successfully demonstrated using a quadrupedal robot.
Moreover, we have compared the proposed method with existing RL \cite{schulman2017proximal} and constrained RL \cite{zhang2022p3o} algorithms. 
The results confirmed the effectiveness of the proposed method, as it was the only method that completed the tasks.
In conclusion, our contributions are threefold:
\begin{itemize}
\item We propose a new reward-shaping process within the CMORL framework, where multiple reward and cost functions are defined stage-wisely.
\item We introduce a practical CMORL algorithm that adapts PPO to handle multiple objectives and constraints.
\item The proposed method has successfully demonstrated various acrobatic tasks, both in simulation and in real-world environments.
\end{itemize}

\section{RELATED WORK}

\subsection{Constrained Multi-Objective Reinforcement Learning}

Multi-objective RL (MORL) is divided into single-policy methods \cite{kyriakis2022pareto, van2013scalarized}, which aim to find one Pareto optimal policy, and multi-policy methods \cite{basaklar2023pdmorl, xu2020prediction, cai2023distributional, abdolmaleki2020distributional, yang2019generalized}, which aim to find a set of Pareto optimal policies. 
Single-policy methods combine multiple objectives into a single objective using utility functions \cite{hayes2022practical} or preference vectors \cite{kyriakis2022pareto, van2013scalarized}, allowing existing RL algorithms to be applied for policy updates. 
Multi-policy methods either simultaneously update policies for multiple preferences \cite{xu2020prediction, abdolmaleki2020distributional} or train a universal policy that can represent a variety of policies by conditioning on preferences \cite{basaklar2023pdmorl, yang2019generalized}. 
Among these, LP3 \cite{huang2022constrained} and CoMOGA \cite{kim2024scale} are CMORL algorithms that extend existing MORL algorithms to consider constraints using Lagrangian \cite{stooke2020responsive} and primal \cite{achiam2017constrained} approaches, respectively. 
The proposed algorithm, CoMOPPO, can be viewed as a single-policy method that simplifies the implementation of CoMOGA.

\subsection{Reinforcement Learning for Legged Robots}

Advances in simulations, such as Isaac Gym \cite{isaacgym}, have made it possible to directly deploy RL policies trained in simulations into real-world environments, significantly increasing efficiency and reducing risk.
Leveraging these advances, sim-to-real techniques, such as terrain curriculum, have enabled quadrupedal robots to successfully navigate challenging terrains, such as slippery surfaces and steep slopes \cite{lee2020learning, ji2022concurrent, margolis2023walk, kumar2021rma, smith2022legged}.
Furthermore, there have been works that enable dynamic parkour-like movements, such as long jumps and two-legged walking, by using novel reward definitions and state representation approaches \cite{cheng2024extreme, zhang2023learning, caluwaerts2023barkour}.
To simplify the definition of reward functions, constrained RL algorithms \cite{altman1999constrained} have been employed for legged robots \cite{kim2024not, lee2023evaluation}. 
These methods exclude safety-related terms, such as body collisions and joint limits, from the reward function and instead use them to define explicit constraints.
These algorithms have demonstrated the effectiveness of constrained RL through robustness to reward weights. 
However, the tasks addressed by them are primarily limited to locomotion.
We expanded these approaches to the CMORL framework to perform tasks requiring more complex acrobatic movements.

\begin{table*}[!tb]
    \centering
    \caption{\textbf{Example of reward and cost definitions for the back-flip task.}
    The reward and cost functions are defined in the first five lines and the following five lines, respectively.
    In the velocity reward, $\mathbf{1}_\mathrm{turn}$ means $1$ if the robot completes a turn and $0$ otherwise. 
    In the balance reward, $\hat{z}_B$ and $\hat{z}_W$ are the $z$-axis unit vectors of the base and world frame, respectively.
    In the style reward, $q^\mathrm{default}_j$ represents the default position of the $j$th joint. 
    In the contact-related costs, $I_C^\mathrm{specified}$ denotes the set of indices of specified links that make contact. 
    During stages other than the jump stage, the foot contact cost outputs the predefined threshold value as the cost is undefined in these phases.}
    \label{tab: example of reward}
    \resizebox{\linewidth}{!}{%
        \begin{tabular}{|ccccccc|}
            \hline
            \multicolumn{1}{|c||}{stage} & \multicolumn{1}{c|}{stand} & \multicolumn{1}{c|}{sit} & \multicolumn{1}{c|}{jump} & \multicolumn{1}{c|}{air} & \multicolumn{1}{c||}{land} & threshold \\ 
            \hline\hline
            \multicolumn{7}{|c|}{reward functions} \\ 
            \hline\hline
            \multicolumn{1}{|c||}{base height} & \multicolumn{1}{c|}{$-|p_z - 0.35|$} & \multicolumn{1}{c|}{$-|p_z - 0.2|$} & \multicolumn{1}{c|}{$\mathbf{1}_{p_z \leq 0.5}\cdot p_z$} & \multicolumn{1}{c|}{$\mathbf{1}_{p_z \leq 0.5}\cdot p_z$} & \multicolumn{1}{c||}{$-|p_z - 0.35|$} & - \\ 
            \hline
            \multicolumn{1}{|c||}{base velocity} & \multicolumn{1}{c|}{$-(v_x^2 + v_y^2 + \omega_z^2)$} & \multicolumn{1}{c|}{$-(v_x^2 + v_y^2 + \omega_z^2)$} & \multicolumn{1}{c|}{$-\mathbf{1}_\mathrm{turn}\cdot \omega_y$} & \multicolumn{1}{c|}{$-s\cdot \omega_y$} & \multicolumn{1}{c||}{$-(v_x^2 + v_y^2 + \omega_z^2)$} & - \\ 
            \hline
            \multicolumn{1}{|c||}{base balance} & \multicolumn{1}{c|}{$-\angle (\hat{z}_B, \hat{z}_W)$} & \multicolumn{1}{c|}{$-\angle (\hat{z}_B, \hat{z}_W)$} & \multicolumn{1}{c|}{$-|\angle (\hat{y}_B, \hat{z}_W)-\pi/2|$} & \multicolumn{1}{c|}{$-|\angle (\hat{y}_B, \hat{z}_W)-\pi/2|$} & \multicolumn{1}{c||}{$-\angle (\hat{z}_B, \hat{z}_W)$} & - \\ 
            \hline
            \multicolumn{1}{|c||}{energy} & \multicolumn{1}{c|}{$-\sum_j \tau_j^2$} & \multicolumn{1}{c|}{$-\sum_j \tau_j^2$} & \multicolumn{1}{c|}{$-\sum_j \tau_j^2$} & \multicolumn{1}{c|}{$-\sum_j \tau_j^2$} & \multicolumn{1}{c||}{$-\sum_j \tau_j^2$} & - \\ 
            \hline
            \multicolumn{1}{|c||}{style} & \multicolumn{1}{c|}{$-\sum_j (q_j-q^\mathrm{default}_j)^2$} & \multicolumn{1}{c|}{$-\sum_j (q_j-q^\mathrm{default}_j)^2$} & \multicolumn{1}{c|}{$-\sum_j (q_j-q^\mathrm{default}_j)^2$} & \multicolumn{1}{c|}{$-\sum_j (q_j-q^\mathrm{default}_j)^2$} & \multicolumn{1}{c||}{$-\sum_j (q_j-q^\mathrm{default}_j)^2$} & - \\ 
            \hline\hline
            \multicolumn{7}{|c|}{cost functions} \\ 
            \hline\hline
            \multicolumn{1}{|c||}{foot contact} & \multicolumn{1}{c|}{-} & \multicolumn{1}{c|}{-} & \multicolumn{1}{c|}{$\mathbf{1}_{|I_C^\mathrm{foot,rear}| = 0}$} & \multicolumn{1}{c|}{-} & \multicolumn{1}{c||}{-} & $0.25$ \\ 
            \hline
            \multicolumn{1}{|c||}{body contact} & \multicolumn{1}{c|}{$\mathbf{1}_{|I_C^\mathrm{body}| > 0}$} & \multicolumn{1}{c|}{$\mathbf{1}_{|I_C^\mathrm{body}| > 0}$} & \multicolumn{1}{c|}{$\mathbf{1}_{|I_C^\mathrm{body}| > 0}$} & \multicolumn{1}{c|}{$\mathbf{1}_{|I_C^\mathrm{body}| > 0}$} & \multicolumn{1}{c||}{$\mathbf{1}_{|I_C^\mathrm{body}| > 0}$} & $0.025$ \\ 
            \hline
            \multicolumn{1}{|c||}{joint position} & \multicolumn{1}{c|}{$\frac{1}{J} \sum_j \mathbf{1}_{q_j > q_j^\mathrm{max}||q_j < q_j^\mathrm{min}}$} & \multicolumn{1}{c|}{$\frac{1}{J} \sum_j \mathbf{1}_{q_j > q_j^\mathrm{max}||q_j < q_j^\mathrm{min}}$} & \multicolumn{1}{c|}{$\frac{1}{J} \sum_j \mathbf{1}_{q_j > q_j^\mathrm{max}||q_j < q_j^\mathrm{min}}$} & \multicolumn{1}{c|}{$\frac{1}{J} \sum_j \mathbf{1}_{q_j > q_j^\mathrm{max}||q_j < q_j^\mathrm{min}}$} & \multicolumn{1}{c||}{$\frac{1}{J} \sum_j \mathbf{1}_{q_j > q_j^\mathrm{max}||q_j < q_j^\mathrm{min}}$} & $0.025$ \\ 
            \hline
            \multicolumn{1}{|c||}{joint velocity} & \multicolumn{1}{c|}{$\frac{1}{J} \sum_j \mathbf{1}_{|\dot{q}_j | > \dot{q}_j^\mathrm{max}}$} & \multicolumn{1}{c|}{$\frac{1}{J} \sum_j \mathbf{1}_{|\dot{q}_j | > \dot{q}_j^\mathrm{max}}$} & \multicolumn{1}{c|}{$\frac{1}{J} \sum_j \mathbf{1}_{|\dot{q}_j | > \dot{q}_j^\mathrm{max}}$} & \multicolumn{1}{c||}{$\frac{1}{J} \sum_j \mathbf{1}_{|\dot{q}_j | > \dot{q}_j^\mathrm{max}}$} & \multicolumn{1}{c||}{$\frac{1}{J} \sum_j \mathbf{1}_{|\dot{q}_j | > \dot{q}_j^\mathrm{max}}$} & $0.025$ \\ 
            \hline
            \multicolumn{1}{|c||}{joint torque} & \multicolumn{1}{c|}{$\frac{1}{J} \sum_j \mathbf{1}_{|\tau_j | > \tau_j^\mathrm{max}}$} & \multicolumn{1}{c|}{$\frac{1}{J} \sum_j \mathbf{1}_{|\tau_j | > \tau_j^\mathrm{max}}$} & \multicolumn{1}{c|}{$\frac{1}{J} \sum_j \mathbf{1}_{|\tau_j | > \tau_j^\mathrm{max}}$} & \multicolumn{1}{c|}{$\frac{1}{J} \sum_j \mathbf{1}_{|\tau_j | > \tau_j^\mathrm{max}}$} & \multicolumn{1}{c||}{$\frac{1}{J} \sum_j \mathbf{1}_{|\tau_j | > \tau_j^\mathrm{max}}$} & $0.025$\\ 
            \hline
        \end{tabular}%
    }
    \vspace{-10pt}
\end{table*}

\section{BACKGROUND}

\subsection{Constrained Multi-Objective Markov Decision Processes}
A constrained multi-objective Markov decision process (CMOMDP) is defined as $\langle S, A, P, \rho, \gamma, R_{1:N}, C_{1:M}\rangle$ with a state space $S$, an action space $A$, a transition model $P$, an initial state distribution $\rho$, a discount factor $\gamma$, $N$ reward functions $R_i(s,a,s')|_{i=1}^N$, and $M$ cost functions $C_j(s,a,s')|_{j=1}^M$. 
A policy is defined as $\pi: S\mapsto \mathcal{P}(A)$, where $\pi(a|s)$ denotes the probability of executing action $a$ in state $s$.
A trajectory is defined as $\tau = \{s_0, a_0, s_1, a_1, ...\}$, where $s_0 \sim \rho$, $a_t \sim \pi(\cdot|s_t)$, and $s_{t+1}\sim P(\cdot|s_t, a_t)$ $\forall t$.
Action value, state value, and advantage functions for the rewards are defined as $Q^\pi_{R_i}(s, a):=\mathbb{E}_{\tau\sim\pi}[\sum_t\gamma^t R_i(s_t,a_t,s_{t+1})]$, where $s_0=s$ and $a_0=a$, $V^\pi_{R_i}(s):=\mathbb{E}_{a\sim\pi(\cdot|s)}[Q^\pi_{R_i}(s, a)]$, and $A^\pi_{R_i}(s, a):= Q^\pi_{R_i}(s, a) - V^\pi_{R_i}(s)$.
Similarly, the value and advantage functions for the costs are defined by substituting $R_i$ with $C_j$.
The reward and cost functions are used to construct objectives and constraints in a CMORL problem, respectively. 

\subsection{CMORL Problem Setup}

A CMORL problem is defined as follows:
\begin{equation}
\begin{aligned}
&\quad \qquad \max_\pi J_{R_i}(\pi) \; \forall i \in \{1, ..., N\} \\
&\mathbf{s.t.} \; J_{C_j}(\pi) \leq d_j/(1-\gamma) \; \forall j \in \{1,...,M\},
\end{aligned}
\end{equation}
where $J_{R_i}(\pi):=\mathbb{E}_{\tau\sim\pi}\left[\sum_t \gamma^t R_i(s_t,a_t, s_{t+1})\right]$, and $d_j$ is a threshold of the $j$th constraint.
The target of the CMORL problem finds a \emph{constrained-Pareto (CP)} optimal policy \cite{kim2024scale}.
Given two feasible policies $\pi_1, \pi_2 \in \{\pi|J_{C_j}(\pi)\leq d_j/(1-\gamma) \; \forall j\}$, if $J_{R_i}(\pi_1)\leq J_{R_i}(\pi_2)$ $\forall i$, $\pi_1$ is \emph{dominated} by $\pi_2$.
If a policy is not dominated by any other policies, the policy is CP optimal.
Given that a CP optimal policy is not unique, a specific policy can be obtained by transforming multiple objectives into a single objective using a \emph{preference vector} $\omega \in \Omega = \{v\in\mathbb{R}^N|\sum_i v_i = 1, \; v_i \geq 0\}$.
This transformation can be achieved either by linearly weight-summing the preference vector and objectives \cite{lu2023multiobjective, hayes2022practical, yang2019generalized} or by using other scalarization methods \cite{van2013scalarized, kyriakis2022pareto}.
Subsequently, single-objective RL algorithms can be applied to obtain the policy.

\section{PROPOSED METHOD}

Now, we introduce a new CMORL framework, which enables intuitive reward shaping for acrobatic tasks. 
The proposed method consists of three main parts: \textbf{1)} stage-wise reward shaping, \textbf{2)} a policy update rule handling multiple objectives and constraints, and \textbf{3)} sim-to-real techniques for deploying policies trained in simulation to real-world.
In the rest of the section, the details of each part will be described.

\begin{figure}[!tb]
    \centering
    \includegraphics[width=0.8\linewidth]{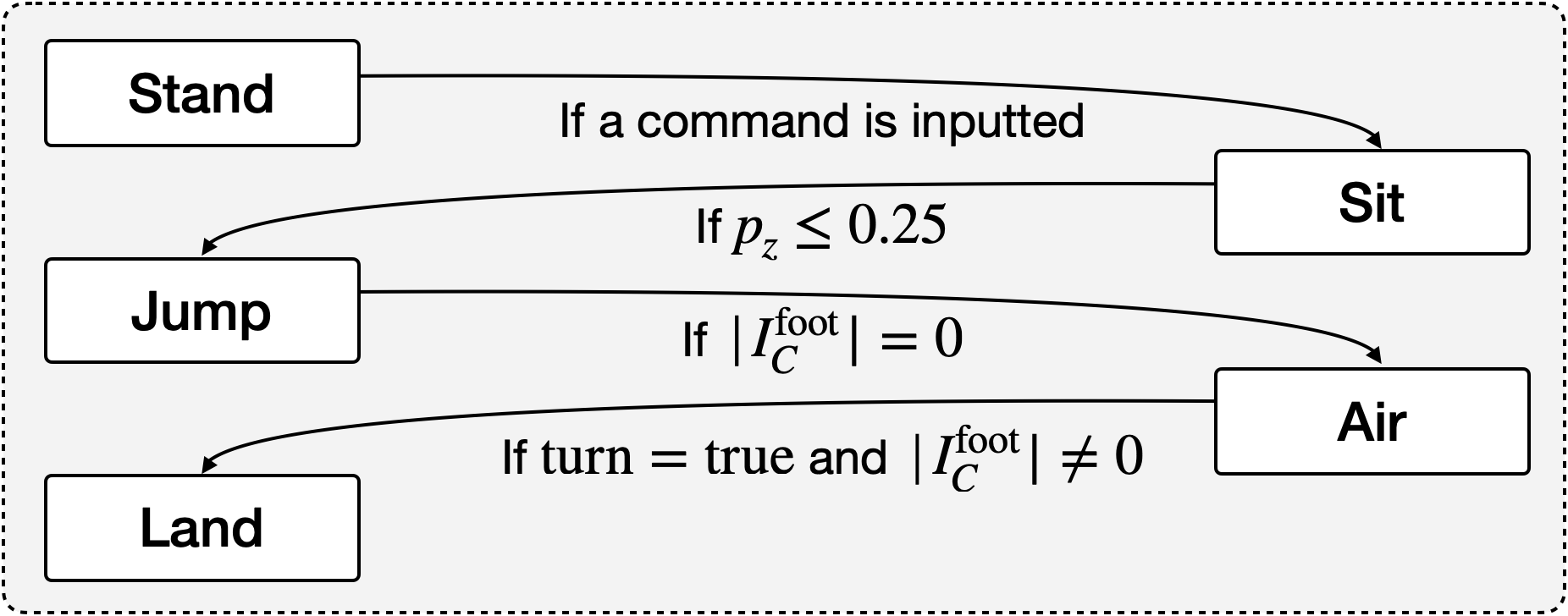}
    \caption{\textbf{Example of stage transitions for the back-flip.} The task start in the stand stage. 
    Upon receiving a back-flip command, the robot attempts to sit. 
    When the base height drops below $0.25\; m$, it transitions to the jump stage. During this stage, the robot attempts to jump, transitioning to the air stage once all feet detach from the ground. After completing the aerial motion, the robot transitions to the landing stage as soon as at least one foot makes contact with the ground.}
    \label{fig: phase}
    \vspace{-10pt}
\end{figure}

\begin{figure*}[!tb]
    \centering
    \begin{subfigure}[b]{1.0\linewidth}
        \centering    
        \includegraphics[width=1\linewidth]{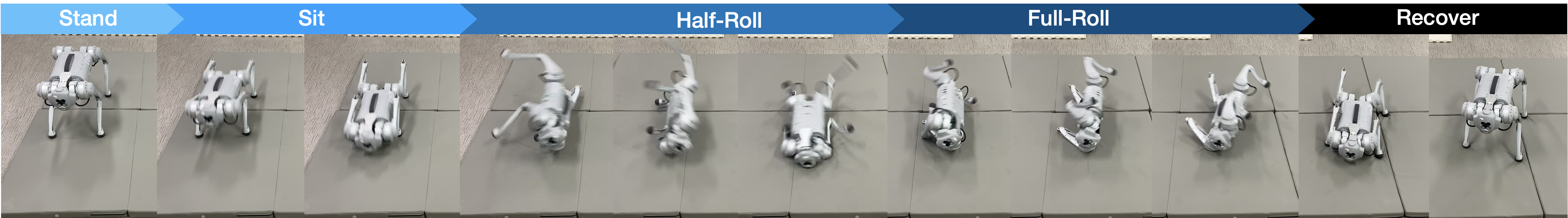}
    \end{subfigure}
    \begin{subfigure}[b]{1.0\linewidth}
        \centering    
        \includegraphics[width=1\linewidth]{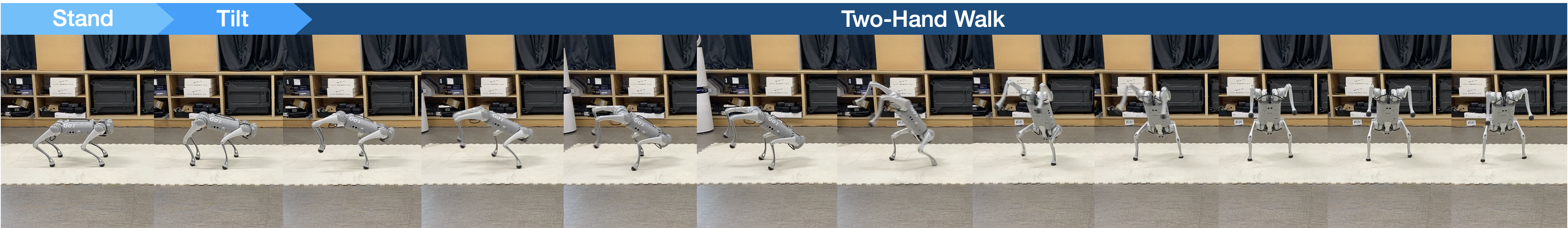}
        \caption{Sim-to-Real Tasks}
    \end{subfigure}
    \begin{subfigure}[b]{1.0\linewidth}
        \centering
        \includegraphics[width=1\linewidth]{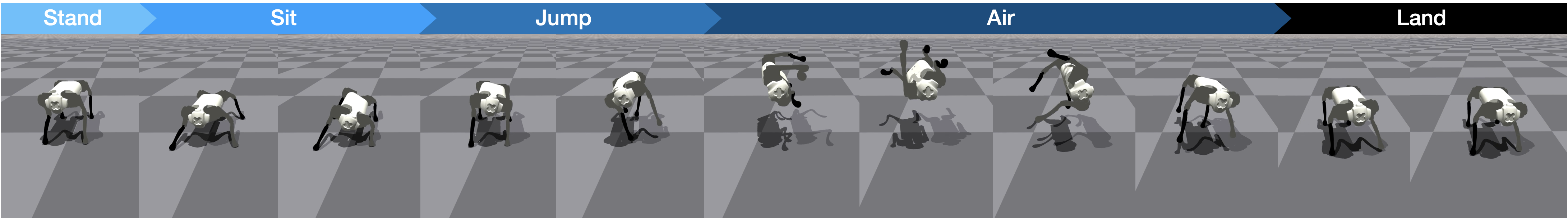}
    \end{subfigure}
    \begin{subfigure}[b]{1.0\linewidth}
        \centering
        \includegraphics[width=1\linewidth]{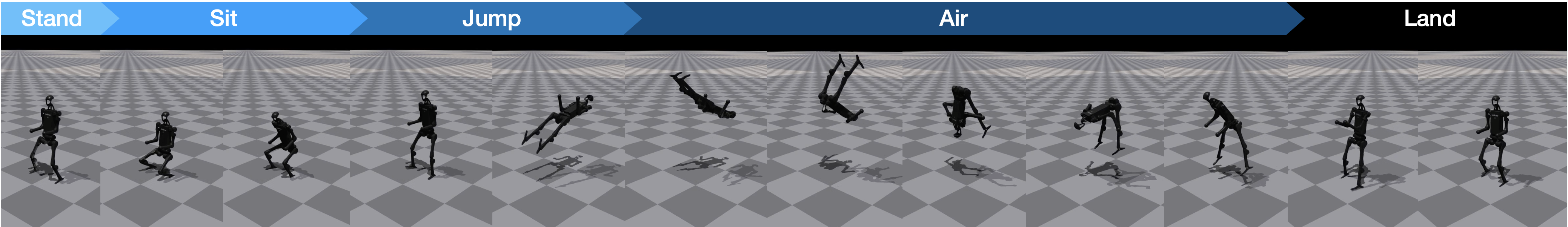}
        \caption{Simulation Tasks}
    \end{subfigure} 
    \caption{\textbf{Snapshots of motion sequences generated by trained policies.} The first two rows show the Unitree Go1 robot performing side-roll and two-hand walk tasks in the real-world environment. The remaining two rows illustrate the Go1 robot performing the side-flip task and the H1 robot performing the back-flip task in simulation.}
    \label{fig: snapshots of results}
    \vspace{-10pt}
\end{figure*}

\subsection{Stage-Wise Reward Shaping}
\label{sec:stage-wise}

In general, a reward function consists of several terms including task-related terms, regularization terms, and safety-related terms.
Instead of integrating all terms into a scalar reward, we use a CMORL framework which maximizes multiple objectives corresponding to each reward term and satisfying constraints corresponding to the safety-related terms.
Furthermore, for a complex task requiring sequential movements, it is required to adjust weights of individual reward terms dynamically, which further increases the difficulty of reward shaping.
To resolve this issue, we propose to divide a task into a sequence of \emph{stages} and define reward and cost functions in a stage-wise manner.
Since segmenting tasks into stages clarifies the required motions for each stage, the reward-shaping process becomes straightforward.
An example of the stages and definitions of the reward and cost functions for the back-flip task are provided in Fig. \ref{fig: phase} and Table \ref{tab: example of reward}, respectively.
Using this example, we provide an overview of how the stages are segmented and how the reward and cost functions are defined.

\subsubsection{Stage Transitions}
As shown in Fig. \ref{fig: snapshots}, to perform a back-flip, the robot should remain in the standby mode until a command is received. 
Once the command is inputted, the robot sits down slightly, then jumps to rotate in the air, and finally lands. 
Through this insight, the task can be divided into five stages named \emph{Stand-Sit-Jump-Air-Land}.

\subsubsection{Reward and Cost Functions}
In the stand, sit, and land stages, the robot is required to remain stationary; therefore, the base velocity reward is defined as the negative of the current velocity.
Conversely, in the jump and air stages, where the robot should rotate backward, the velocity reward is set to the $y$-directional angular velocity. 
Also, to ensure sufficient jump height during these stages, the height reward is set to the base height.
In order to prevent tilting during the jump and air stages, the balance reward is set to remain the angle between the $y$-axis of the base and the $z$-axis of the world perpendicular. 
In the other stages, the reward is set to minimize the angle between the $z$-axis of the base and the world frames to maintain the robot upright.
The energy and style rewards are used as regularization to ensure natural motions. 
The body contact cost prevents the robot from falling over, while costs associated with joint position, velocity, and torque are implemented to limit those values within their respective ranges.
The foot contact cost is defined specifically for jumping with the rear legs; the cost is incurred if the rear legs detach before the front legs.

\subsection{CMORL  Policy Update}
\label{sec: policy update}

In this section, we introduce a policy update rule, termed \emph{constrained multi-objective PPO (CoMOPPO)}, designed to maximize multiple objectives while satisfying constraints.
According to Kim et al. \cite{kim2024scale}, convergence to a CP optimal policy can be achieved by aggregating the advantage functions of rewards and costs through a weighted summation, where the weights satisfy specific conditions, and updating the policy using TRPO \cite{schulman2015trust} with the aggregated advantage.
It can be written as follows:
\begin{equation}
\label{eq: CoMOGA update}
\begin{aligned}
\pi_{t+1} = &\; \underset{\pi}{\mathrm{argmax}} \underset{\tau\sim \pi_{t}}{\mathbb{E}} \left[ \frac{\pi(a|s)}{\pi_{t}(a|s)} A^{\pi_{t}}(s,a) \right] \\
&\mathbf{s.t.} \; \mathbb{E}_{\tau\sim\pi_t} \left[ D_\mathrm{KL}(\pi_{t}(\cdot|s)||\pi(\cdot|s)) \right] \leq \epsilon,
\end{aligned}
\end{equation}
where $A^{\pi_{t}}(s, a) := \sum_i \nu_{t,i} A_{R_i}^{\pi_{t}}(s,a) - \sum_j \lambda_{t,j} A_{C_j}^{\pi_{t}}(s,a)$, $\epsilon$ is a trust region size, and $D_\mathrm{KL}$ is the KL divergence.
For the condition on the weights, $\nu$ and $\lambda$, please refer to Theorem 4.2 in \cite{kim2024scale}.

In order to properly determine the weights, $\nu$ and $\lambda$, we use \textbf{1)} reward normalization and \textbf{2)} standard deviation of advantage functions.
First, in the CMORL setting, each reward function operates at a different scale, making it essential to adjust them to a consistent level.
To this end, we apply reward normalization for each reward and stage, and train the value functions using the normalized rewards. This approach automatically adjusts the ratio of each objective to a consistent level.
Next, it is important to maintain consistency not only in the scale of the rewards but also in the ratio between objectives and constraints.
Without a consistent ratio, the policy may be updated to over-maximizing objectives rather than satisfying constraints, potentially destabilizing the training process.
To resolve this, we normalize the reward and cost advantages by their respective standard deviations, ensuring the policy to be updated with a consistent ratio of objectives to constraints.
As a result, the proposed rule for advantage aggregation as follows:
\begin{equation}
A^\pi = \frac{A_R^\pi}{\mathrm{Std}[A_R^\pi]} - \eta \sum_j \frac{A_{C_j}^\pi}{\mathrm{Std}[A_{C_j}^\pi]} \mathbf{1}_{(J_{C_i}(\pi)>d_i)},
\end{equation}
where $\mathrm{Std}$ denotes the standard deviation, $A_R^\pi := \sum_i \omega_i \hat{A}_{R_i}^\pi$ with a given preference $\omega$, $\hat{A}_{R_i}^\pi$ is the advantage function calculated from the normalized rewards, and the hyper-parameter $\eta$ serves as the ratio of constraints, as done in \cite{zhang2022p3o}.
With the aggregated advantage functions, the policy can be updated using \eqref{eq: CoMOGA update}.
However, to simplify the implementation, we apply a PPO update, which is formulated as follows \cite{schulman2017proximal}:
\begin{equation*}
\pi_{t+1} \!\!= \underset{\pi}{\mathrm{argmax}} \! \underset{\tau\sim \pi_{t}}{\mathbb{E}} \! \left[\min(r_t A^{\pi_t}, \mathrm{clip}(r_t, 1-\epsilon, 1+\epsilon)A^{\pi_t}) \right],
\end{equation*}
where $r_t:=\pi(a|s)/\pi_t(a|s)$, and $\epsilon$ is a hyper-parameter.

\begin{figure*}[!t]
    \centering
    \begin{subfigure}[b]{0.325\linewidth}
        \centering    
        \includegraphics[width=1\linewidth]{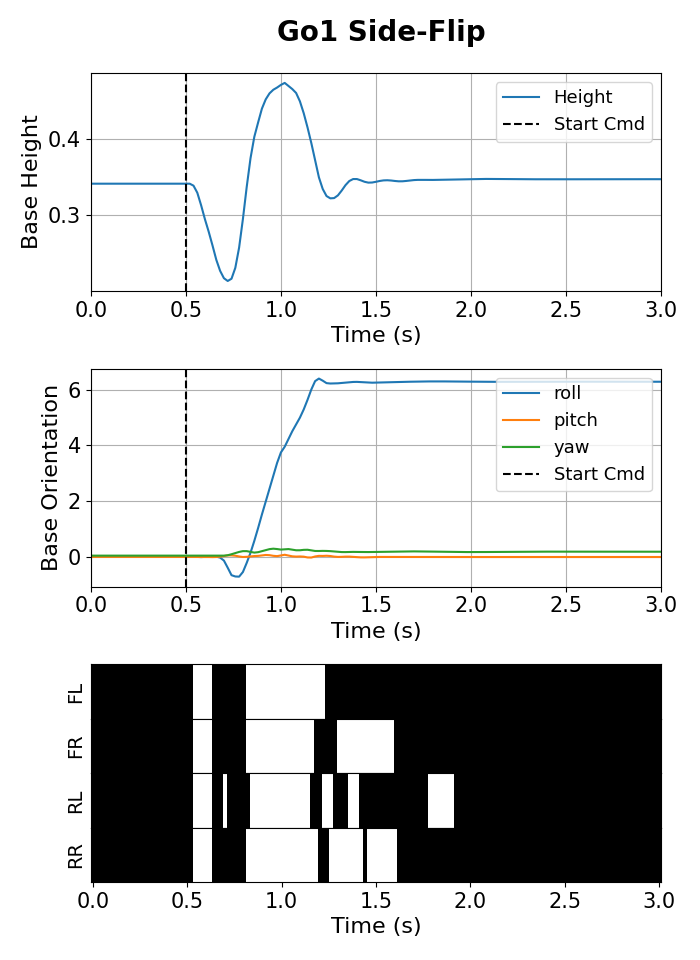}
    \end{subfigure}
    \hfill
    \begin{subfigure}[b]{0.325\linewidth}
        \centering    
        \includegraphics[width=1\linewidth]{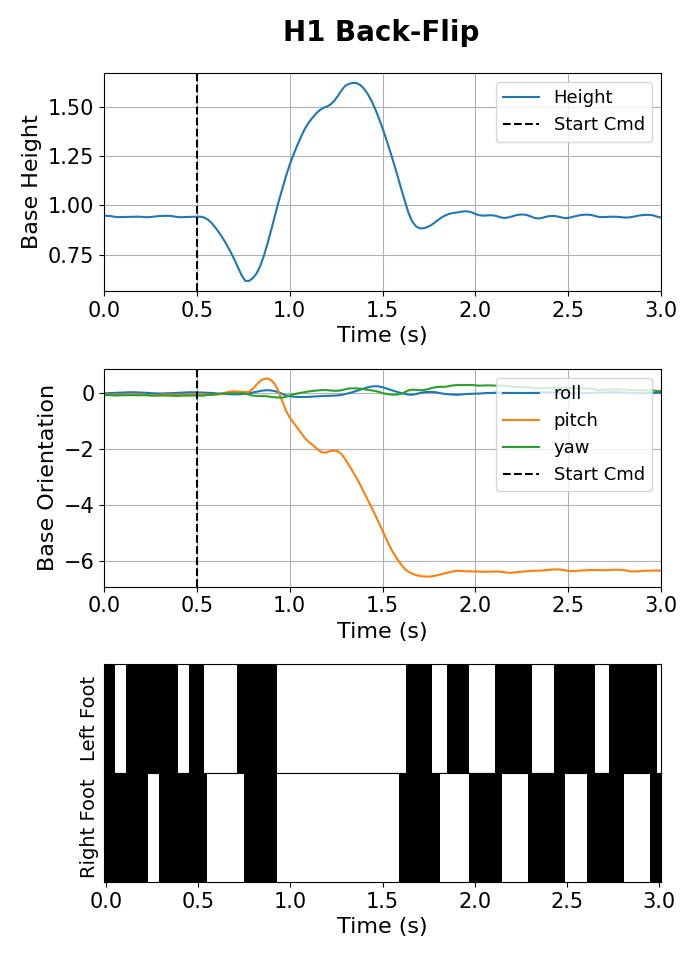}
    \end{subfigure}
    \hfill
    \begin{subfigure}[b]{0.325\linewidth}
        \centering    
        \includegraphics[width=1\linewidth]{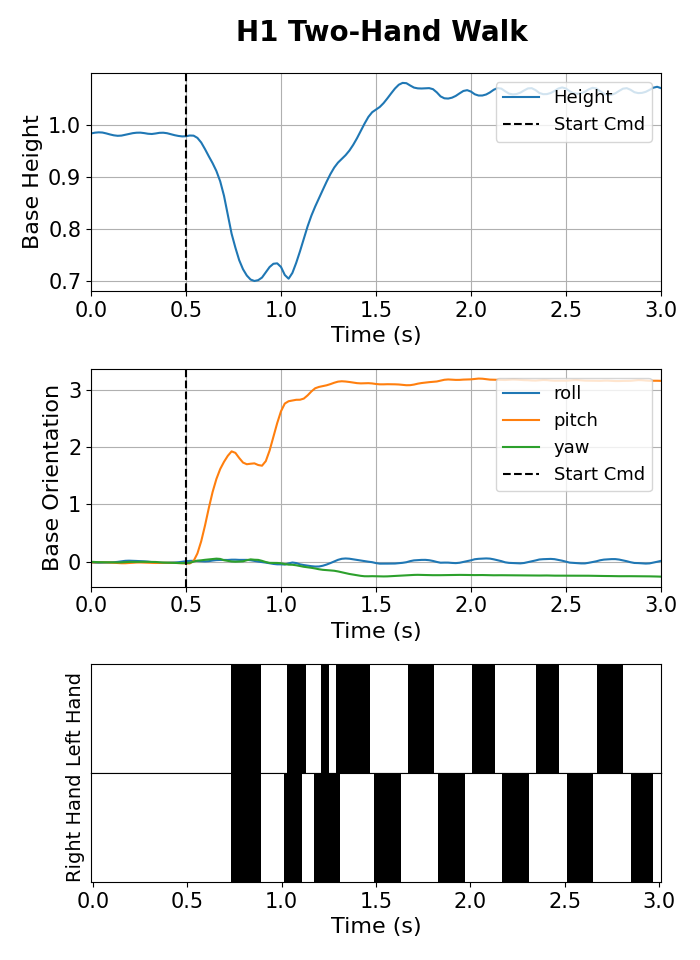}
    \end{subfigure}
    \vspace{-5pt}
    \caption{\textbf{Simulation experiment results.} 
    The first two rows show the changes in height and body orientation, while the last indicates whether the specified parts of the robot are in contact with the ground (black) or not (white) in each task.
    }
    \label{fig: sim task results}
    \vspace{-10pt}
\end{figure*}

\subsection{Sim-to-Real Techniques}
\label{sec:sim-to-real}

In order to deploy policies trained in simulation to real-world environments, we use two widely-used sim-to-real techniques: \textbf{1)} domain randomization and \textbf{2)} teacher-student learning \cite{lee2020learning, kim2024not, agarwal2023legged}. 
For domain randomization, we follow an RL approach proposed in \cite{margolis2023walk}, which involves randomizing motor strength and offset, gravity, friction, restitution, noise in joint positions and velocities, and base orientation. 
Details on the range of randomization are provided in the appendix of \cite{margolis2023walk}.
Next, teacher-student learning \cite{lee2020learning} is used to distill a teacher policy, which uses states including privileged information, into a student policy that relies solely on sensor data, such as joint position and base orientation.
The process is similar to the method proposed in \cite{lee2020learning}, but the action execution differs slightly.
In our approach, the actions of the teacher and student policies are alternately provided to the environment at fixed intervals.
This helps in collecting a dataset for teacher-student learning that closely aligns with the student policy.

\section{EXPERIMENTS}

This section describes the tasks and their corresponding results for both simulation and real-world environments.
Details on the motions generated by the trained policies, along with the corresponding reward and cost functions for each task, can be found in the attached video.

\subsection{Environmental Setup}

We use the Isaac Gym simulator \cite{isaacgym} due to its effectiveness in sim-to-real transfer and its flexibility in creating a wide range of tasks across various robotic platforms.
In simulation experiments, we employ two types of robots from Unitree Robotics: Go1, a quadrupedal robot, and H1, a humanoid \cite{unitree}. 
The quadrupedal robot comprises 23 body links and 12 motors, while the humanoid robot features 20 body links and 19 motors. 
In real-world experiments, we deploy the quadruped robot, Go1, through sim-to-real techniques as discussed in Section \ref{sec:sim-to-real}.

The state representation includes body orientation, joint positions and velocities, commands, as well as the previous action.
For the teacher policy, privileged information—such as linear and angular velocities, height, and foot contact, which are difficult to access in the real-world but available in simulation—is additionally used.

\begin{figure*}[!t]
    \centering
    \begin{subfigure}[b]{0.335\linewidth}
        \centering    
        \includegraphics[width=1\linewidth]{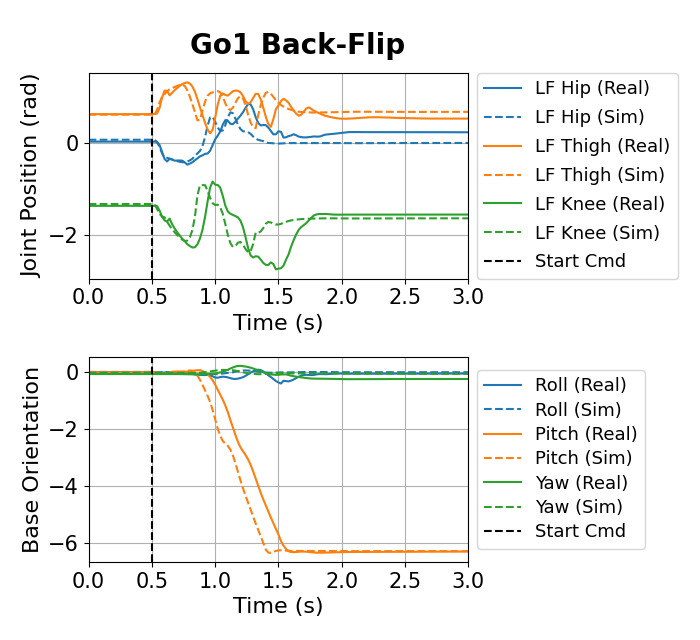}
    \end{subfigure}
    \hspace{-12pt}
    \begin{subfigure}[b]{0.335\linewidth}
        \centering    
        \includegraphics[width=1\linewidth]{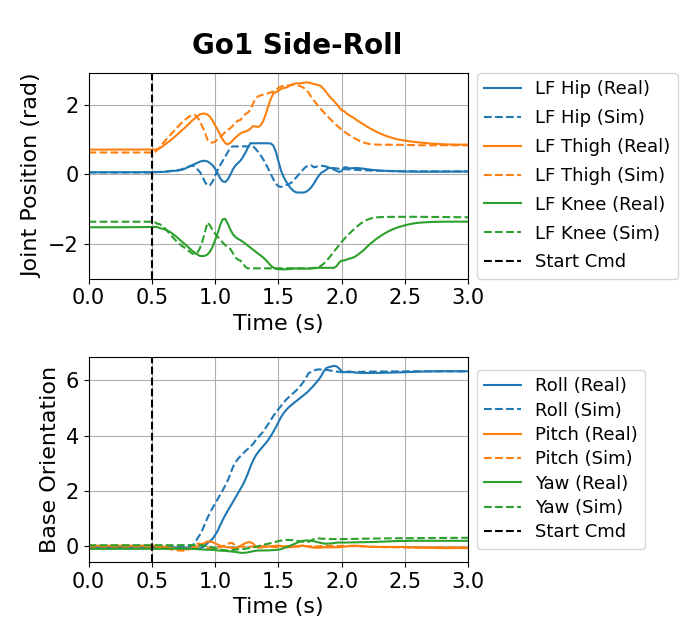}
    \end{subfigure}
    \hspace{-12pt}
    \begin{subfigure}[b]{0.335\linewidth}
        \centering    
        \includegraphics[width=1\linewidth]{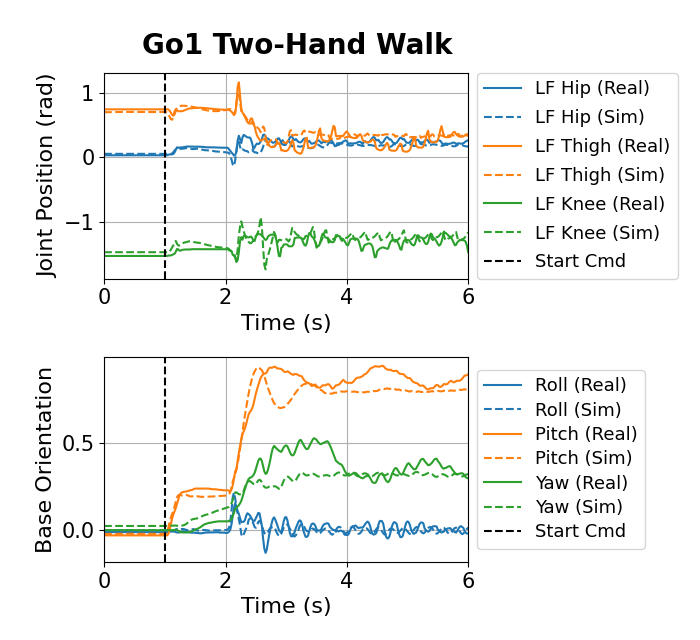}
    \end{subfigure}
    \vspace{-5pt}
    \caption{\textbf{Sim-to-real experimental results.} 
    The graph shows the position changes of three joints—hip, thigh, and knee—in the left front leg, along with the body orientation over time for each task. 
    The solid line represents the real-world data, while the dotted line indicates the simulation results.
    }
    \label{fig: real-world task results}
    \vspace{-5pt}
\end{figure*}

\begin{figure*}[!t]
    \centering
    \includegraphics[width=1\linewidth]{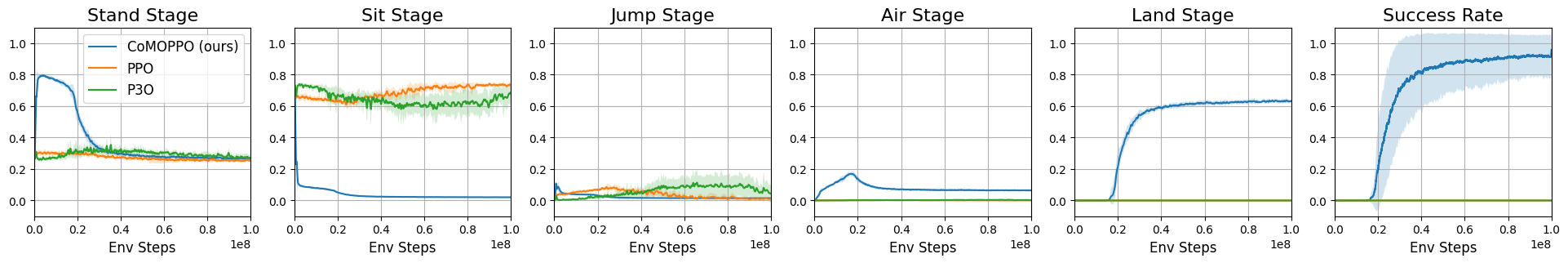}
    \caption{\textbf{Comparison with RL and constrained RL algorithms.} 
    The first five graphs show the proportion of time the robot remains in each stage during an episode, while the last shows the success rate. 
    All results are obtained by training algorithms with five random seeds.
    Bold lines represent the mean, and shaded areas indicate the standard deviation of the results.
    }
    \label{fig: ablation}
    \vspace{-10pt}
\end{figure*}

\subsection{Task Details}

We have designed four acrobatic tasks: \emph{back-flip, side-flip, side-roll}, and \emph{two-hand walk}. 
The back-flip and two-hand walk tasks are implemented on both types of robots, while the side-roll and side-flip tasks are designated for the quadruped.
Snapshots of each task are shown in Fig. \ref{fig: snapshots} and Fig. \ref{fig: snapshots of results}, and the following are descriptions of each task.
 
\subsubsection{Back-Flip}
The robot jumps backward into the air, rotates 360 degrees without touching the ground, and lands on its feet in its initial pose.
The stage transition, along with the reward and cost functions, is discussed in Section \ref{sec:stage-wise}.

\subsubsection{Side-Flip}
This task is similar to the back-flip, but the robot jumps to the right side instead of backward. 
The stage transitions remain identical to those of the back-flip.

\subsubsection{Side-Roll}
The robot performs a full roll along its right side, returning to its original pose upon completion.
This task is segmented into five stages: \emph{Stand}, where the robot remains upright; \emph{Sit}, where the robot lowers itself to prepare for the roll; \emph{Half-roll}, where the robot lies on its back; \emph{Full-roll}, where the robot completes the roll; and \emph{Recover}, where the robot returns to its default pose and orientation.

\subsubsection{Two-Hand Walk}
The robot performs walking using its hands or front legs, while maintaining balance.
This task is divided into three stages: \emph{Stand}, where the robot maintains its default pose; \emph{Tilt}, where the robot lowers its front legs or places its hands on the ground to prepare for standing; and \emph{Walk}, where the robot walks using only its two hands.

\subsection{Results}

As illustrated in Fig. \ref{fig: snapshots} and Fig. \ref{fig: snapshots of results}, the robots were able to successfully execute the tasks in both simulation and real-world.
In simulation, the robot precisely performed the required maneuvers, returning to its original pose without losing stability during the side-roll and flip tasks.
As shown in Fig. \ref{fig: sim task results}, it is clearly demonstrated that during the Go1 side-flip task, the robot first lowered its height and rotated to the right side, as indicated by the roll angle.
In the H1 back-flip task, the change in the pitch angle and the height indicates that the robot flipped backward.
For the H1 two-hand task, the agent exhibited stable balance, consistently alternating its support between the two hands.

Similarly, in the sim-to-real experiments, the robot was able to replicate the learned behaviors with high precision.
To provide a more detailed comparison between the simulation and real-world performance, we analyzed the changes in the joint positions of the left front leg and body orientation throughout the tasks.
As shown in Fig. \ref{fig: real-world task results}, the robot successfully jumped backward in both simulation and real-world.
In the side-roll task, the change in base orientation followed a similar pattern to the side-flip task, as the robot moved sideways in both tasks.
In the two-hand walk, the robot was tilted forward, as shown by the pitch angle, while oscillations in the joint positions imply that the robot was repeatedly lifting and placing its feet to stabilize its balance.

The robot's real-world behavior closely matched that observed in the simulation, displaying similar patterns in joint position and orientation changes. 
This strong resemblance confirms the success of sim-to-real process, demonstrating that the trained policies were effectively transferred from the simulation to the real environment.
The actual movements of the robot during these tasks can be seen in the attached video.

\subsection{Ablation Study}

To demonstrate the efficacy of CMORL, we compared the proposed method with existing RL and constrained RL algorithms in the Go1 back-flip task.
In the case of RL, we employed PPO \cite{schulman2017proximal}, where a single reward function is defined by weight-summing the multiple reward and cost functions, and the weights are obtained from the average ratios of objectives and constraints calculated during the training of CoMOPPO.
For constrained RL, we utilized penalized PPO (P3O) \cite{zhang2022p3o}, where a single reward function is defined by summing the multiple rewards with the same weights used in PPO, and the constraints are the same as in CoMOPPO.

As illustrated in Fig. \ref{fig: ablation}, CoMOPPO was the only algorithm that successfully complete the task.
In contrast to CoMOPPO, which transitions quickly to the \emph{Air} stage after brief \emph{Sit} and \emph{Jump} stages, the other two algorithms remained primarily in the \emph{Sit} stage, indicating that they were unable to properly execute the jump motion.

\section{CONCLUSIONS}

In this work, we have proposed an RL method that defines reward and cost functions in a stage-wise manner within the CMORL framework \cite{kim2024scale}.
Additionally, we have developed a practical CMORL algorithm by expanding PPO \cite{schulman2017proximal} to handle multiple objectives and constraints.
The proposed method has successfully demonstrated acrobatic tasks in both simulation and real-world settings. 
Moreover, by comparing the proposed method with existing RL and constrained RL algorithms, we have shown the necessity of the CMORL framework. 
While in this work, tasks are manually segmented into stages, future research could investigate more efficient techniques for automatically dividing complex tasks into stages.

\bibliographystyle{IEEEtran}
\bibliography{main}

\addtolength{\textheight}{-12cm}   



\end{document}